# Multi-Objective Optimization of Water Resource Allocation for Groundwater Recharge and Surface Runoff Management in Watershed Systems

Abbas Sharifi[1], Hajar Kazemi Naeini[2], Mohsen Ahmadi[3], Saeed Asadi[2], Abbas Varmaghani[4]

[1]Department of Civil and Environmental Engineering, Florida International University, Miami, FL, USA
[2]Department of Civil Engineering, University of Texas at Arlington, Arlington, Texas, USA
[3]Department of Electrical and Computer Science, Florida Atlantic University, Boca Raton, FL, USA
[4]Department of Computer Engineering, Islamic Azad University of Hamadan, Hamedan, Iran
Corresponding author: mahmadi2021@fau.edu

**Abstract**

Land degradation and air pollution are primarily caused by the salinization of soil and desertification that occurs from the drying of salinity lakes and the release of dust into the atmosphere because of their dried bottom. The complete drying up of a lake has caused a community environmental catastrophe. In this study, we presented an optimization problem to determine the total surface runoff to maintain the level of salinity lake (Urmia Lake). The proposed process has two key stages: identifying the influential factors in determining the lake water level using sensitivity analysis approaches based upon historical data and optimizing the effective variable to stabilize the lake water level under changing design variables. Based upon the Sobol'-Jansen and Morris techniques, the groundwater level and total surface runoff flow are highly effective with nonlinear and interacting impacts of the lake water level. As a result of the sensitivity analysis, we found that it may be possible to effectively manage lake levels by adjusting total surface runoff. We used genetic algorithms, non-linear optimization, and pattern search techniques to solve the optimization problem. Furthermore, the lake level constraint is established based on a pattern as a constant number every month. In order to maintain a consistent pattern of lake levels, it is necessary to increase surface runoff by approximately 8.7 times during filling season. It is necessary to increase this quantity by 33.5 times during the draining season. In the future, the results may serve as a guide for the rehabilitation of the lake.

***Keywords:*** Revitalization, Optimization, Sensitivity Analysis, Runoff control, Groundwater.

# 1. Introduction

Urmia lake is a high salinity lake in northwestern Iran with considerable volume and water level fluctuations. Water levels have dropped dramatically in recent decades due to a deficiency of river system input and excess outflow due to evaporation. River systems inflow is affected by human involvement and climate change, whereas evaporation-based outflow is more affected by global warming and water level (Lahijani et al. 2021). Salt accumulation spreads throughout the lake's bottom and coastline during the significant water level drop. Considering the absence of water intake, the bottom salt load produces a drop in basin volume, which modifies basin physiography in terms of water level. A total of $2,790 \times 10^6$ m$^3$ of salt has been accumulated during the previous few decades, equating to a 73-centimeter-thick salt coating on the lake's bottom. Since this salt load plays a critical role in filling the basin volume, the water volume of Urmia lake cannot be obtained using the basin volume-water level relation during extreme water level in fall (Lahijani et al. 2021). The lake's salinity has grown to more than 300 g/l in recent years as a result of drought and rising agricultural water needs in the lake's basin, and huge parts of the lake bottom have been dried (Eimanifar and Mohebbi 2007). Lakes drying up, among other things, can have a huge influence on food supply. At the basin level, lakes integrate climatic variability. These changes have an impact on food production and consumption (metabolism) in lake environments, both nearshore and in open water. While the shallow nearshore ecosystem is vital for agricultural production, many lake studies concentrate on the offshore ecosystem (Scordo et al. 2022). Lake drying might make it difficult to provide the consistent, inexpensive, and nutritious food supplies needed to keep the world's population healthy (Feizizadeh et al. 2022). Food consumption and production were lower in dry and wet years than in years with medium meteorological conditions. Additionally, throughout dry and wet years, animal biomass dropped, and the nearshore bottom ecosystem was more vulnerable to changes due to a considerable fall in food supply. Differential influences within lake ecosystems are caused by climatic variables. As a result, if only one habitat is researched, lake management may be inadequate (Scordo et al. 2022, Nikpour, et al. 2025, Reihanifar et al. 2024 ).

Hydrologic parameters are important for preserving a water body's structure and function, and they influence a variety of abiotic elements, which might have an impact on the biota that grows there. Since saline lakes are found largely in endorheic basins, their size, salinity, and annual mixing regimes fluctuate with changes in their hydrologic budgets, they may be vulnerable to environmental changes (Romero and Melack 1996). The lake's mining, along with additional salinity from irrigation, has resulted in an increase in soil salinity in the basin, leading to desertification(Fathian et al. 2015). These environmental changes may have severe consequences such as the spread of illnesses, devastation of agricultural fields, and huge economic harm, leading to widespread exodus of local people, similar to what has occurred in the Aral Sea in recent decades (AghaKouchak et al. 2015). The efficiency of water demand management strategies, especially in limiting agricultural water use increases, is strongly dependent on regular implementation and supervision (Bakhshianlamouki et al. 2020, Asadi et al. 2025). Lake desiccation, salinization, and desertification in the basin have been caused by misguided agricultural practices and climate change. Desertification and salinization have spread over the basin. The concerns increasing desertification and air pollution in northwest Iran are soil salinization and desertification owing to the drying up of Urmia lake and dust produced from the parched bottom of the lake (Alizade Govarchin Ghale et al. 2019).

The lake's full dryness will surely cause an environmental disaster in the region. Desiccation of the lake would result in the formation of a desert and saline body covering around 6000 $km^2$, which will be a source of sandstorms and air pollution in the vicinity. The Aral Sea is a wonderful illustration of how to comprehend and evaluate the causes and consequences of Urmia Lake's drying up (Alizade Govarchin Ghale et al. 2019). Increased evaporation and misuse of surface water and groundwater resources have accompanied unsustainable agricultural expansion over the Urmia Lake basin. Before developing an implementation strategy, it is essential to have a solid knowledge of the system in the form of a unified conceptual model that has been validated against many field data types (Danesh-Yazdi and Ataie-Ashtiani 2019). The hydro-chemical relationship between Urmia Lake and groundwater remains contentious and incompletely known. In fact, the absence of information on water stable isotopes or chemical tracers makes it hard to distinguish mixing processes, flow patterns, residency lengths, and water source not just in the Urmia lake basin, but across the area (Shamsi and Kazemi 2014; Danesh-Yazdi and Ataie-Ashtiani 2019). In this study, an optimization problem was developed to determine the total annual surface runoff flow that might be required to maintain the level of Urmia Lake. It involves two stages: identifying effective factors in the lake water level through sensitivity analysis based on historical data and optimizing the most important variable to stabilize the lake water level with changing design variables. First, we develop a meta-model based on a deep neural network method in order to identify the relevant factors. In order to identify the most significant input parameters, we utilized the Sobol'-Jansen and Morris methods. Second, the input variables are rearranged in such a manner as to make the prior stage's effective variables dependent on the second stage's input variables. Based on results of the sensitivity analysis, the most effective element is considered the output layer in the meta-model, and the rest of the variables are considered the input layer. In order to optimize the system, three approaches are employed: genetic algorithms (GAs), pattern searches, and nonlinear programming.

## 2. Literature review

Maleki et al. (2022) investigated the disappearance of Urmia Lake using climate-smart agriculture practices. Nonetheless, the potential advantages, possibilities, dangers, costs, and incentive mechanisms of the interventions have not been thoroughly explored. This qualitative study was conducted to determine the most effective intervention for the lake's restoration and to look at the major development factors. The Fuzzy VIKOR technique simplified the ranking of climate-smart agriculture solutions and indicated the best practicable strategy for reducing the Urmia Lake disaster in the eastern half of the basin based on the findings. The hydrogeological state of the route for the water transmission tunnel to Urmia Lake was assessed by Amiri and Asgari-Nejad (2022). Geophysical analysis, borehole drilling, and pressure variations at various depths of aquifers utilizing composite piezometers, LeFranc permeability test, and pumping test were used to obtain data. They proposed that water be carried to the lake by a tunnel from the Kanisib Dam owing to the area's morphology and geography, as well as the transmission route. Sorkhabi et al. (2022) used wavelet decomposition and a convolutional neural network to study waveform retracking in Urmia Lake from 1992 to 2019. The proposed approach improved the accuracy of waveform retracking by up to 30%. The climate experiment and gravity recovery, as well as the yearly monitoring of the water level station, were utilized to verify satellite contour maps, with substantial correlations of 0.66 and 0.96, respectively, with satellite altimetry. Ahmadi et al. (2024)

explored the application of Digital Twin models combined with deep learning for terrain segmentation in coastal areas. Using USGS data and satellite imagery, they classified landscapes into seven categories with high accuracy, employing a U-Net-based segmentation model. Their findings emphasize the role of Digital Twins in effectively monitoring landform changes and coastal environmental conditions. Moghim et al. (2023) analyzed extreme hydrometeorological events in Bangladesh by employing the WRF model to simulate intense rainfall and cyclone effects. They improved rainfall forecasts using a Bayesian regression approach and introduced a flash flood index (FFI) to pinpoint vulnerable regions. Their study provides insights for designing resilient infrastructure and developing effective flood mitigation strategies based on future climate projections. Hussain et al. (2024) analyzed meteorological drought in Ankara, Turkey, using logistic regression models. The RELogRM model was identified as the best for predicting fall-to-winter drought transitions. Their findings show that higher spring moisture reduces summer drought risk by 24.16%, emphasizing the role of antecedent conditions in drought forecasting.

KhazaiPoul et al. (2019) used a mix of soil and water assessment models to investigate the tradeoff between agricultural productivity and environmental water demand. The results revealed that by optimizing reservoir operation through irrigation management, water stress indicators may be reduced from 80% in traditional irrigation patterns to 60%, with only a 6.5 percent reduction in agricultural output. Tarebari et al. (2018) investigated several facets of sustainable water resource management, such as economic, social, and environmental factors, while also attempting to address stakeholder disputes using non-symmetric Nash bargaining, which is connected to the multi-objective optimization technique. The findings revealed that the number of available resources or reservoir volume has a massive effect on the optimal degree of the utility function and the efficiency of the proposed method, with the higher number of resources or larger reservoirs resulting in a higher optimal degree of the utilitarian calculus. For simulating the restoration possibilities of Urmia Lake based on the prey-predator strategy. Barhagh et al. (2021) adopted a system dynamics technique. Increasing the lake water level by 2.54 m, 2.04 m, 1.19 m, and 1.19 m, respectively, through lowering the number of irrigated fields by 40%, boosting irrigation efficiency, reducing the lake area itself, and inter-basin water transfer. For selecting water management techniques, Dehghanipour et al. (2020) used a simulation-optimization methodology. Multiple solutions, such as combining minimal environmental flow needs, deficit irrigation, and crop selection, are identified as increasing environmental flow (up to 16 percent) and agricultural profit (up to 24 percent) at the same time when compared to historical conditions. Danandeh Mehr et al. (2023) developed VMD-GP, an advanced model for drought prediction in ungauged catchments. Using GDM data, VMD, and GP, it accurately forecasts SPEI and outperforms traditional models, improving drought monitoring.

To obtain the best tuning settings for lake water level simulation, Ehteram et al. (2021) used the sunflower optimization technique. Different water harvesting possibilities were studied, taking into account environmental constraints and equitable water distribution to stakeholders. According to the findings, examining Urmia Lake water harvesting scenarios revealed that a 30% water harvesting scenario enhances the lake's water level. On the Urmia lake analysis, researchers used optimization strategies such as Extremely powerful learning machine (Sales et al. 2021), Firefly Algorithm (Valipour et al. 2019), Fruit Fly Optimization (Samadianfard et al. 2019), Bayesian Belief Networks (Dehghanipour et al. 2021), Rainfall Network (Valipoor et al. 2020), Wavelet-cuckoo search (Komasi et al. 2018), ANN-whale optimization (Vaheddoost et al. 2020), Particle swarm (Gamshadzaei and Rahimzadegan 2021). Several datasets were created for the Urmia Lake. Based on Nasseri et al. (2022) spatiotemporal calibration architecture, eight global high-resolution monthly precipitation datasets such as CHIRPS, ERA-5, IMERG, PERSIANN, PERSIANN-CCS, PERSIANN-CDR, TRMM, and Terra were created. According to their observations, the original IMERG dataset overstated monthly precipitation by around 20% when compared to rain gauge data.

## 3. Methods and Material

### 3.1. Artificial Neural network

An input layer, hidden layers of neurons, and output neuron make up an artificial neural network. A multi-layer perceptron (MLP) design with lines linking neurons. Each link has a weight, which is a numerical value. The output, $h_i$, of neuron $i$ in the hidden layer is,

$$H_j = A\left(\sum_{i=1}^{n} w_{ij} x_i + b_j^{hid}\right) \quad (1)$$

Let $A$ as a activations function (such as Sigmoid, Tangent hyperbolic) , $n$ is the number of of input features, $w_{ij}$ the weights matrix, $x_i$ is the vector of to the input layer data, and $b_j^{hid}$ the bias value of the hidden layer. In addition to updating nonlinearity into the neural network, the A function is responsible for limiting the neuron's value to prevent divergent neurons from paralyzing it. To define a neural network, factors such as interconnections, layer count, training algorithm, propagation rules, etc. must be specified. There are two aspects to consider when it comes to the MLP includes the training stage and the predicting process. In both phases, the number of layers and activation functions must be same.

### 3.2. Sobol'–Jansen sensitivity analysis

For sensitivity analysis, the Sobol'–Jansen approach performs Monte Carlo to estimate of the Sobol' indices (Nossent et al. 2011). It decomposes the output variance using the summation of input variable variances to expand the numerical field's dimension. To establish the magnitude of the fluctuations in the model output, this approach employed random values for the parameters. This form of breakdown is analogous to the traditional statistical analysis of variance. Assume f is the model and x is the input variable; we now have:

$$y = f(x) = f(x_1, \ldots, x_n) \tag{2}$$

As a result, the total unconditional variance is calculated as follows:

$$Var(y) = \sum_j V_j + \sum_j \sum_{m>j} V_{jm} + \cdots + V_{l,\ldots,k} \tag{3}$$

$$V_i = Var_{x_i}(E x \sim i(y|x_i)) \tag{4}$$

$$V_{ij} = Var_{x_{ij}}(E x \sim ij(y|x_i, x_j) - V_i - V_j \tag{5}$$

The $x_i$ notation denotes the set of all variables except $x_i$, where $V_j$ and $V_{jm}$ are the first- and second-order variance terms, respectively. As a result, the Sobol'–Jansen technique may be used to determine $S_j$ (first-order sensitivity) and $S_j^T$ (total sensitivity):

$$S_j = \frac{V_j}{Var(y)}, S_j^T = \sum_{k \neq j} S_K \quad (6)$$

The high value of $S_j$ indicates that an input variable has a large impact on the output without considering probable interactions between the input parameters. Furthermore, if the output's uncertainty is minimized, $S_j^T$ can be used to assess sensitivity. Because the uncertainty of $x_j$ has no effect on the uncertainty of the output model y when $S_j$ is modest, $x_j$ is inferred as a non-influential or inconsequential variable. Whether $S_j^T$ is large or little, $x_j$ has no effect on other variables. As a result, the tiny values of $S_j$ and $S_j^T$ indicate that $x_j$ uncertainty is unrelated to y uncertainty.

### 3.3. Morris' sensitivity analysis

Morris' approach is a sensitivity analysis tool that is quick and easy to use. This method, also known as the elementary effect method, allows for the identification of the model's basic variables, which might include components that interact. When the trials only examine one element at a time and the number of variables is big enough to need computationally expensive simulations, this approach can be used. Morris is a local approach in which the output of the model is generated by altering the value of only one input between two simulation runs. At each level, the local measure of incremental ratios like an elementary effect is determined, accounting for the whole input space. The measurements are averaged to provide a single effect that is independent of sample point selection. Morris(Morris 1991) presented two sensitivity measures based on a basic impact to assess if the influence of xi on y is minimal, linear, and additive, or non-linear and nonadditive. The terms "elementary effect" and "measurement" are used interchangeably.(Balesdent et al. 2016):

$$R_i = \frac{f(x_1, \ldots, x_{i-1}, x_i, x_{i+1}, \ldots, x_n) - f(x_1, \ldots, x_n)}{\Delta} \quad (7)$$

The n-dimensional input space discretized into a p-level grid $\Omega$ is defined by n independent inputs $x_i, i = 1, 2, \ldots, n$.

The distribution of elementary effects $F_i$ for each ith input is generated by randomly picking a vector $X$ from the grid $\Omega$ on input space. The sensitivity measures of the distribution $F_i$ are the mean $\mu_i$ and standard deviation $\sigma_i$ estimations. $\sigma_i$ reflects the cumulative effects of input owing to nonlinearity and interactions with other inputs, whereas mean $\mu_i$ evaluates the total influence of input on the output. The second sensitivity measure, $\sigma$, can be used to analyze factors that interact with other parameters or have non-linear effects, such as the standard deviation of the distribution of incremental ratios (Saltelli et al. 2008). The presence of a big $\mu$ value for input indicates that the input has an influence on the output. Large $\sigma$ values, on the other hand, imply that the input parameter has a non-linear influence on the model output or interacts with one or more other parameters. To decide which metric is more important, we plot $\mu$ (spread) and $\sigma$ (strength). The solution would then be the parameters in the upper right corner of the figure, where both sensitivity measurements are high.

### 3.4. Constrained optimization algorithms

In data science, constrained optimization challenges are common. Artificial intelligence is an example of this. Let $\{x_1, x_2, \ldots, x_n\}: \mathbb{R}^n$, the objective is to extremize a given measure while keeping the predicted parameters below or equal to a specific threshold value. As a result, this problem may be expressed as:

Extremize $y = f(x_1, x_2, \ldots, x_n) \in \mathbb{R}^n$ (8)

s.t. $x_i \leq c$ (9)

Where $f$ is an n-dimensional pre-trained function and $x_i$ is the input factors or function evaluation. The restricted optimization approach extremizes the y value if $x_i$ is less than or equal to c, which is a constant number. The principle of survival of the fittest underpins the genetic algorithm (GA) method to optimization. The GA is an evolutionary algorithm since it mimics the processes of evolution. The most important aspects get stronger as a result of this process, while the weaker elements are removed. A stochastic search of the solution space utilizing strings of binary variables known as chromosomes, which represent the parameters being optimized, is used to solve an optimization problem using the GA approach. Each cell inside these chromosomes is known as a gene, and each gene has a binary value for these modeling purposes (Murray-Smith 2012). To get the relevant parameters, a random population of chromosomes is produced and decoded. The system model is then updated with these parameter values. Using a performance index based on a fitness function, a simulation is conducted, and results are obtained for each set of parameters within the population. When the fitness values are discovered, they are sorted into ascending order with the chromosomes that belong to them. The most important fitness values are picked and then exposed to crossover and mutation processes(Murray-Smith 2012; Wilson and Mantooth 2013). The crossover operator generates two new chromosomes (offspring), and the process is continued until enough offspring have been produced to replace the 80 percent of the current population with the lowest cost values. The mutation also entails the random selection of a particular number of genes in the present population, followed by random changes in their values. To arrive at a final answer, the technique is repeated for a predetermined number of iterations (generations) (McCall 2005).

Pattern search algorithms are a type of direct search algorithm that looks for the minimizer of a continuous function without utilizing derivatives. The well-known approach of Hooke and Jeeves (Hooke and Jeeves 1961) and the simplex algorithm of Nelder and Mead(Nelder and Mead 1965) were among the earliest direct search algorithms. These were regarded heuristics at the time, with no formal convergence theory. The objective function is assessed on a finite collection of nearby points on a carefully designed discrete mesh at each iteration of the Pattern search, which is produced by examining nonnegative integer combinations of vectors that constitute a positive spanning set (Lewis and Torczon 1996). If the new iteration improves the mesh, it is accepted and the mesh is kept or coarsened; if not, the mesh is refined, and a new set of surrounding mesh points is assessed.

### 3.5. Conceptual model of the present work

This research presented an optimization problem for determining the possible surface runoff flow or river discharge needed to maintain the level of Urmia Lake on an annual basis. Figure 1 depicts the conceptual diagrams of the suggested technique. The process consists of two primary stages: discovering effective factors in the lake water level using sensitivity analysis approaches based on historical data and optimizing the most important variables to stabilize the lake water level with changing design variables. The river's outflow is the most compelling element of the lake water level, according to the sensitivity analysis results. The outcomes will be shown in the Results section. To determine the influential variables, we first present a meta-model based on a deep neural network method.

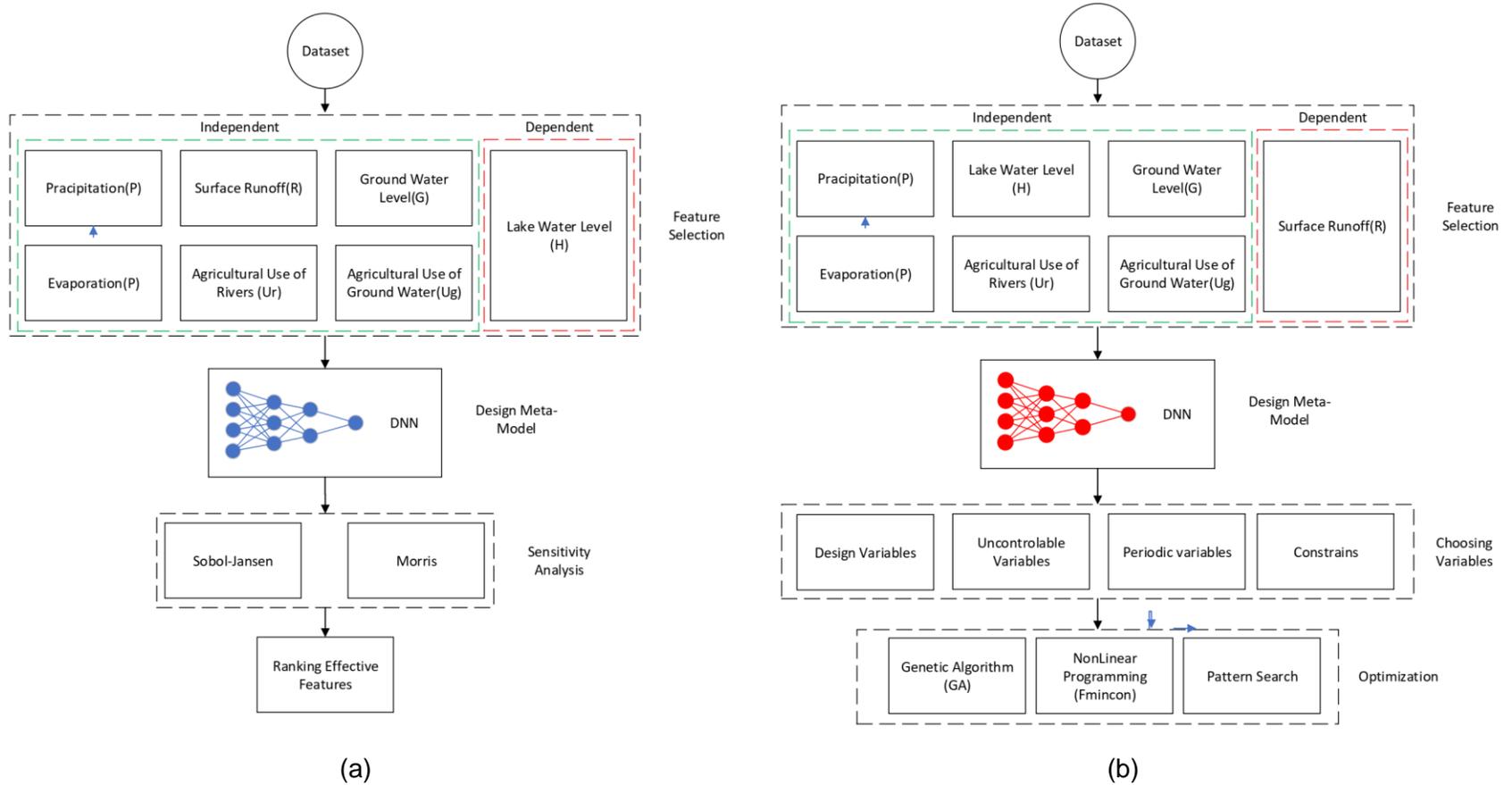

**Fig. 1** Conceptual diagram of the provided method, a) Diagram of sensitivity analysis for choosing controlling influential variables, b) Diagram of optimization based on objective variable provided by Fig. 2 (a)

Standardized variables in the [0, 1] interval is used to train the meta-model. The input parameters were then ranked using two sensitivity analysis approaches, such as Sobol-Jansen and Morris. These methods rely on the generation of random points in homogeneous variable spaces. To decompose the dependent variance, the Sobol-Jansen approach uses the summing of variances of independent variables to increase the dimension of the numerical field. To establish the magnitude of the fluctuations in the model output, this approach employed random values for the parameters. Furthermore, the Morris technique computes the model output by altering the value of only one input between successive simulation runs. At each level, the local measure of incremental ratios like an elementary effect is determined, accounting for the whole input space.

As a result, the variables are ranked according to how successful the input variables are in influencing the dependent characteristic. The input variables are rearranged in the second step so that the influential variables from the first stage are deemed dependent. We aim to define an objective function to stabilize the water level as one of the design variables based on this arrangement. A meta-model based on deep neural networks and optimization algorithms are included in this step, which is referred to as the optimization stage. Total Surface runoff (R) is regarded the output layer in the meta-model based on the sensitivity analysis results, while the rest of the variables are employed as input layers. The goal of this optimization problem is to maximize the value of surface runoff in order to maintain a steady water level based on a yearly pattern. In the optimization process, we should pick periodic, uncontrollable, and constraint design variables. As a result, three approaches are used to carry out the optimization: genetic algorithm (GA), pattern search, and non-linear programming (Fmincon). The results are also documented in order to determine the potential value of surface runoffs and the pattern of consistent water levels.

## 4. Results and Discussion

### 4.1. Data Collection

Urmia Lake is the second largest inland Salt Lake in the world. It is supplied with water by 17 major rivers, 12 seasonal rivers, and 39 floodways. Several rivers and streams drain into this lake, passing through agricultural and urban areas, as well as industrial and industrial zones. Figure 2 illustrates the river system of the lake and the location of wells for using ground water for agricultural purposes. The variables used to estimate the water level are listed in Table 1. As the lake's surface area has decreased, the elevation of its surface has declined.

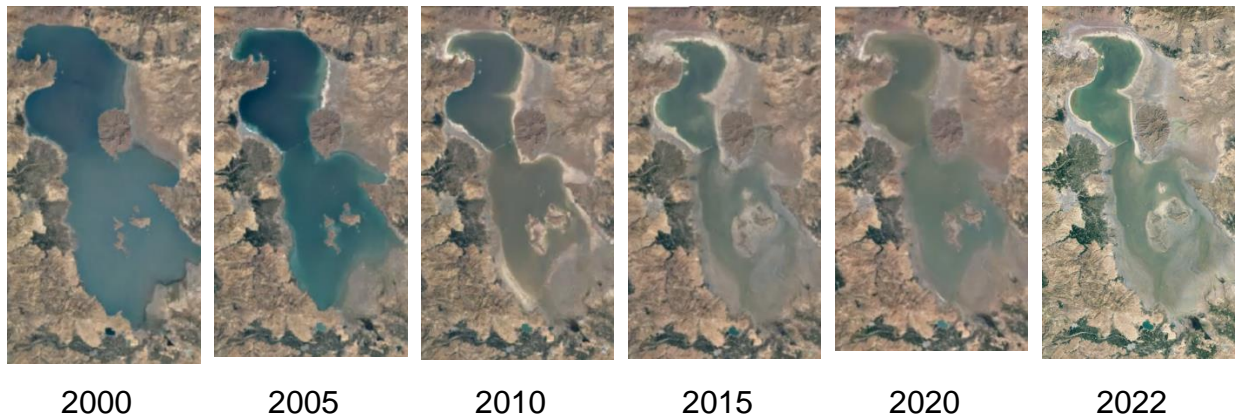

2000     2005     2010     2015     2020     2022

(a)

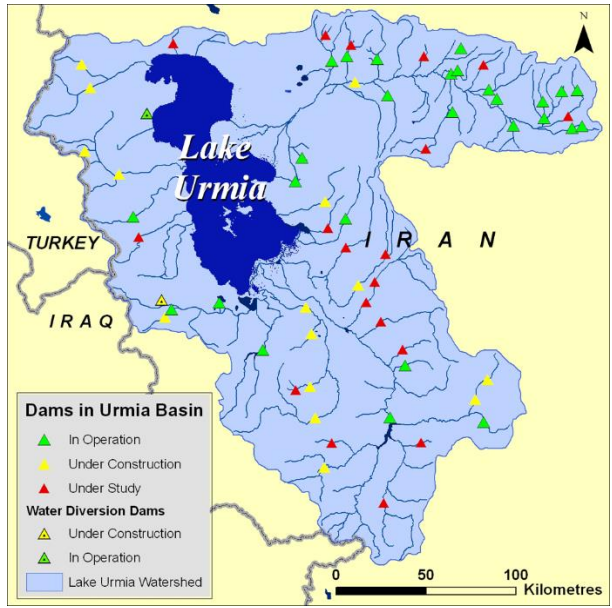
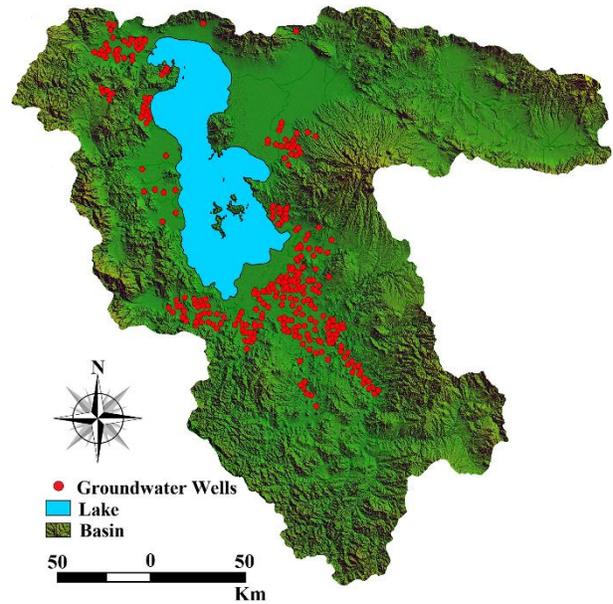

(b)             (c)

**Fig. 2** a) Historical map of Urmia Lake from the last 22 years b) The river system and dams in the Urmia Lake basin, s) The location of wells in the basin

**Table 1:** The governing features for sensitivity analysis and optimization problem, IN: Independent, D: Dependent, The term above mean sea level (a.m.s.l)

| Variable | Unit | Type | Properties | Dependency | | Min | Max | Mean | STD |
|---|---|---|---|---|---|---|---|---|---|
| | | | | Sensitivity Analysis | Optimization | | | | |
| Precipitation (P) | mm | Historical | Precipitation on lake Area | IN | IN | 0.00 | 173.00 | 25.31 | 29.86 |
| Total Surface Runoff (R) | m$^3$/s | Historical | Total Surface Inflow to Lake from rivers | IN | D | 0.10 | 686.78 | 59.72 | 98.92 |
| Groundwater Level (G) | m a.m.s.l | Historical | Average Groundwater Level in the Adjacent Aquifers of Lake | IN | IN | 1295.5 | 1298.8 | 1297.2 | 0.848 |
| Evaporation (E) | mm | Historical | Evaporation from lake based on Remotely Sensed Data | IN | IN | 0.36 | 293.23 | 95.27 | 84.12 |
| Agricultural Use of Rivers (Ur) | mm | Periodic pattern | Computed monthly time-averaged root-zone water balance components for agricultural zones within the Miyandoab aquifer boundary | IN | IN | 18.86 | 241.64 | 103.71 | 74.37 |
| Agricultural Use of Groundwater (Ug) | mm | Periodic pattern | | IN | IN | 0.00 | 46.52 | 17.93 | 18.45 |
| The lake water level (H) | m a.m.s.l | Historical | Urmia lake Water Level | D | IN | 1270.0 | 1274.6 | 1272.0 | 1.34 |
| The lake water level constraint pattern (Hcon) | m a.m.s.l | Periodic pattern | Periodic pattern of the Lake Water Level based on 2018 as reference | Constrain value | Constrain value | 1270.3 | 1270.8 | 1270.5 | 0.205 |

As a result of reduced river flow to the lake, excessive evaporation, and below-average precipitation, the lake's surface area has decreased. The rate of evaporation in saline lakes, such as Urmia Lake, is influenced by salt concentration and climate conditions. Since the construction of the causeway, the salinity gradient between the northern and southern portions of the lake has increased. High volumes of precipitation that falls on the lake and river discharges increase the lake evaporation rate by decreasing salinity and expanding the lake's surface area. A lake evaporation model is required to accurately forecast the spatiotemporal variability of evaporation while accounting for the impacts of lake water volume and salt content (Danesh-Yazdi and Ataie-Ashtiani 2019).

Agriculture is also the principal source of water imbalance in the Urmia Lake Basin. The major consequence of irrigation mismanagement, in addition to the reduction in water flow to the lake, is the following soil salinization due to inadequate water irrigation and soil erosion due to over-cultivation, which puts too much stress on an already stressed ecosystem. Sustainable irrigation water management should meet two goals at once: protecting the natural environment while supporting irrigated agriculture for food security. Improving irrigation methods would be a good way to relieve stress on Urmia Lake and its environment while also ensuring regional prosperity. The E is calculated using in situ observations and remote sensing data in this investigation. Simultaneously, the values of the components H, P, and R are derived from in-situ measurements. Based on the Urmia lake Restoration Program, all hydrometeorological ground data utilized in this work was obtained from (Parizi et al. 2022). The monthly precipitation (P) falling directly on the lake water surface was determined using two-gauge stations on the lake's western and eastern edges, Urmia and Tabriz. The sum of monthly runoff from the 12 permanent rivers that flow into the lake was used to compute R (Sheibani et al. 2020). Furthermore, the values of Ug and Ur are calculated using the findings of (Dehghanipour et al. 2019), which are based on monthly time-averaged root-zone water balance for agricultural zones inside the Miyandoab aquifer border.

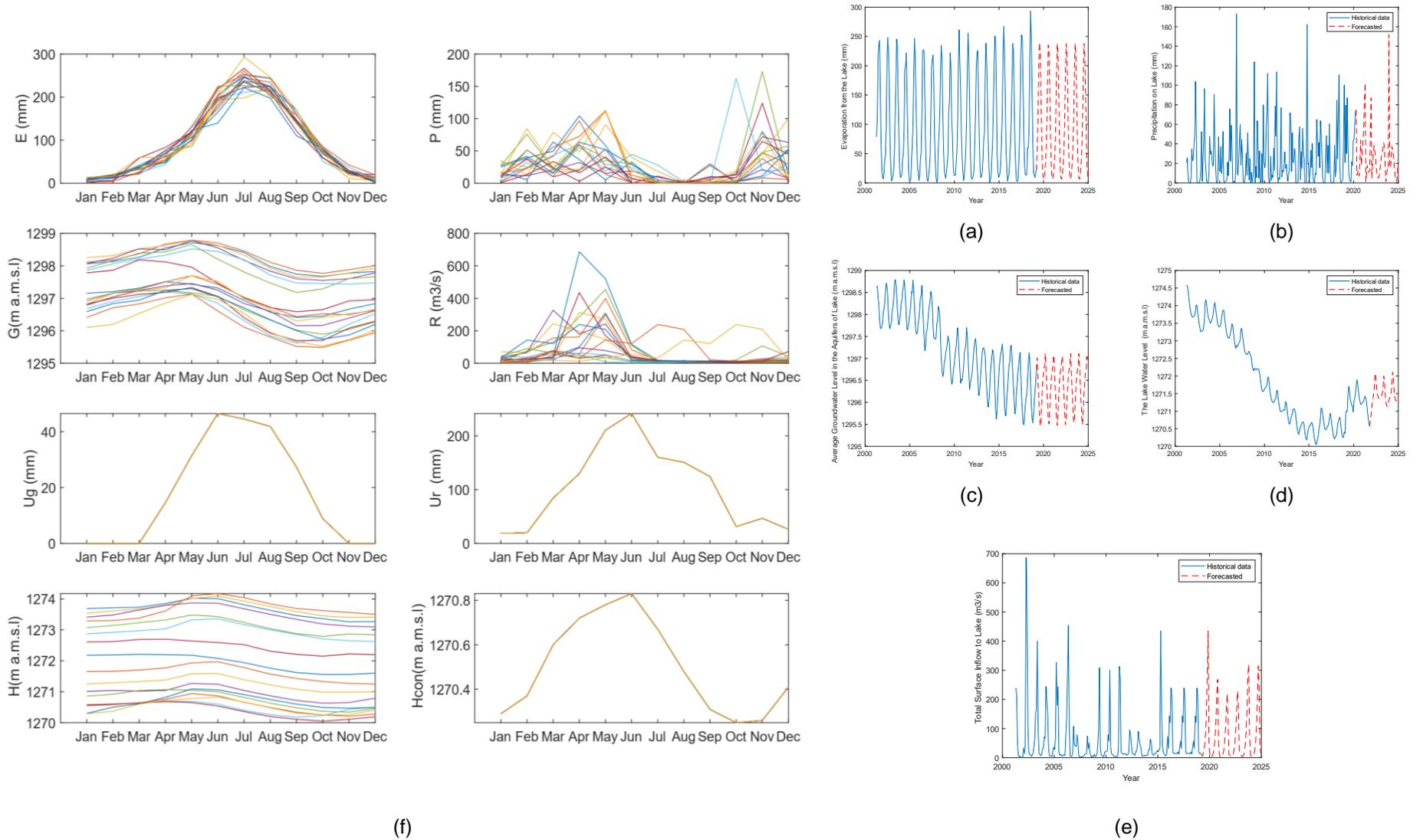

(f)

**Fig. 3** Forecasted value of the features until 2025 using the LSTM method, and Monthly value of the features in the study time interval

Figure 3 depicts the variables' historical values from 2001 to 2019. The key objective for selecting the time frame mentioned is to develop a plan based on historical facts before implementing Iran's National Restoration Program. We employed the Long-Short-Term-Memory (LSTM) approach as one of the machine learning methods to anticipate the characteristics involved in this challenge till 2025 in order to obtain the viewpoint of the governing variables.

For all variables, the forecaster $R^2$ for this problem is about 99%. The monthly value of the variables is shown in Figure 3. According to the monthly evaporation values from the lake, the highest points of evaporation occur during the summer months of June, July, and August. In addition, the amount of precipitation varies from season to season during the spring and fall. In spite of uniform fluctuations in evaporation, precipitation and total surface runoff to the lake vary from year to year despite uniform fluctuations in evaporation. We can identify a consistent pattern to manage the water level in the lake when it comes to river outflow to the lake. Furthermore, the highest ground water level occurs towards the end of the spring and the minimum occurs in the middle of the fall. This fluctuation pertains to the end of each year's agricultural usage of groundwater. Moreover, two yearly pattern is used as agricultural pattern of using ground water and river runoffs for agricultural uses. this pattern is extracted based on the results of the (Dehghanipour et al. 2019) monthly time-averaged of Miyandoab aquifer between 2002-2013. Also, the maximum and minimum value of the lake water level belongs to June and November, respectively. Based on the *H* and *Hcon* value from November to June, the water level increased and from June to November reached the minimum value. Regarding Figure 3, it can be seen that all the variables except evaporation decrease in the summer and fall.

### 4.2. Results of Designing meta-models

According to Kolmogorov's theorem, a three-layer multilayer perceptron may exactly define any continuous function. For the estimate to be precise, nevertheless, an activation function and parameters for which no computation procedures exist must be established. It's impossible to employ any activation function since we need to consider the functions' goal of simplifying the problem. Figure 1 depicts a concept for solving an optimization problem. To train a neural network, we first gathered a dataset in the domain of the variables. Data is standardized in [0,1] intervals in order to develop a robust architecture for sensitivity analysis. Three hidden layers with 30, 20, and 10 neutrons each make up the structure of the given neural network topologies. We also utilized a Logarithmic sigmoid for hidden layers and a Linear function for output layer approximation when activating the model. The first model (Model I) is used for sensitivity analysis, while the second (Model II) is employed to solve optimization problems.

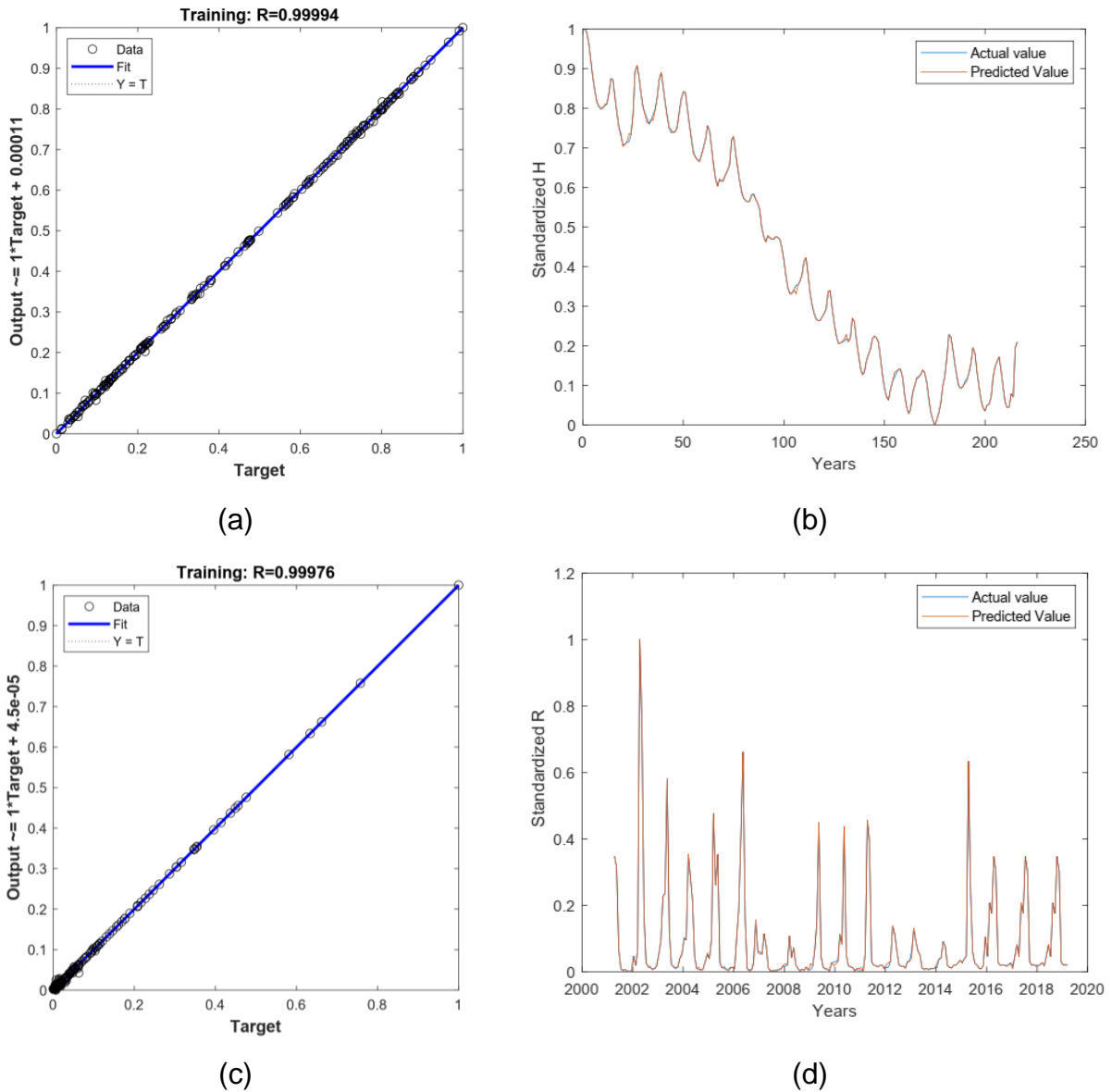

**Fig. 4** Results of the presented meta-model using DNN (a) and (c): Scatter plot of the model I and II respectively, (b), (d), Fitting plot of the model I and II respectively.

To solve the optimization problem, the multilayer perceptron is used to describe the optimization problem's objective function as a nonparametric black box. Finally, design variables are changed to maximize the goal function Figure 4 shows the results of designing meta-models for sensitivity analysis (Model I) and optimization problem (Model II). The presented models are trained using standardized features based on a Deep Neural network. Regarding the results, the $R^2$ of both models is almost 100%, fitting in the fluctuated dependent variable. Consequently, the Model I will be used for sensitivity analysis and finding the influential variables of lake water level as dependent variables. Furthermore, Model II will be used to optimize the surface runoff to stabilize the lake water level.

The provided meta-models are 7-dimensional models with the best fitting of the historical data on the model. To illustrate the 3D-response surface of the presented DNN models, we plot the surface of three variables by fixing other ones on the mean value. Figures 5 a, b, and c show the 3D surface of Model I, and Figures 5 d, e, and f depict Model II. In this regard, other standardized parameters have a constant value of 0.5. Regarding Fig.8 a, the maximum value of the lake level occurred when both Precipitation and River runoff were in the maximum value. Also, based on Fig. 8 b, the maximum water level in the lake occurs when we have the maximum groundwater level. Moreover, with the lowest evaporation value, the maximum value of water in the lake can be observed. Based on the agricultural pattern of the using groundwater and river flow in lands, it can be found that when we have a higher water level in the lake, agricultural use of water is also high. Figure 5 f, g, and h are also plotted based on Model II, considering surface runoff as a dependent variable. Figure 5 is just a 3D surface to show the matching of the historical facts and surface variation. The data for the variables based on the Kolmogorov–Smirnov test is not based on a normal distribution, and each parameter interact to the others.

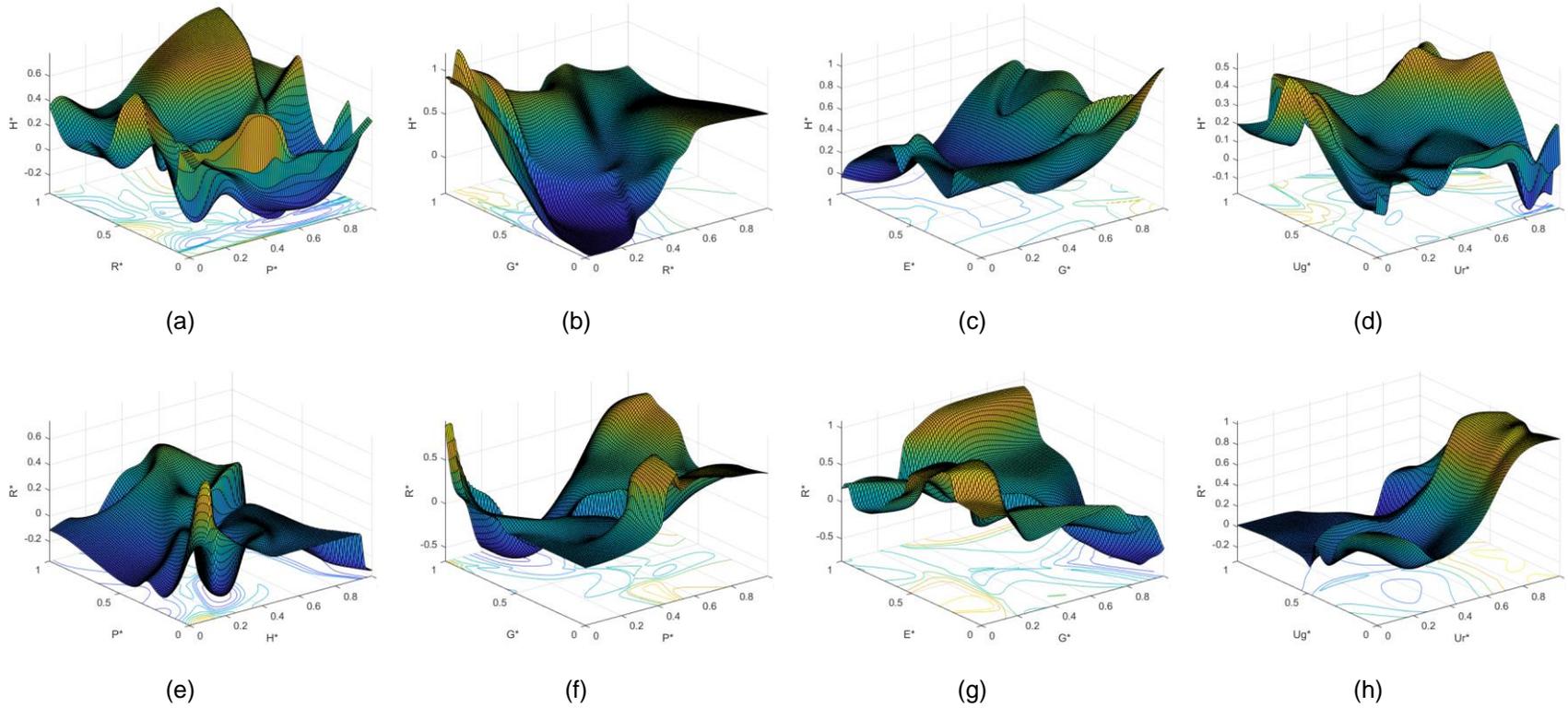

**Fig. 5** 3D-surface of the meta-models based on the DNN approach, a, b, c, d belong to Model I, and e, f, g, h belong to the Model II

Because the system is non-linear, we cannot deduce that changing one parameter causes other changes based on the surface plot. Therefore, DNN, as a meta model, is the best and most efficient choice based on this fact. The given model may estimate the dependent variable with modifying other characteristics because normality of data is not required for DNN. When creating nonparametric models, no explicit assumptions about the functional form are made, unlike when creating parametric models. Nonparametric models, on the other hand, may be thought of as a function approximation that comes as near to the data points as feasible. Nonparametric models have the benefit over parametric techniques in that they may accurately match a greater variety of possible forms for the real function by removing the assumption of a certain functional form, such as a linear model. As a result, we employed the Model I function offered as a nonparametric sensitivity analysis function.

### 4.3. Sensitivity Analysis

First order and total indices' can be calculated concurrently using the Sobol'–Jansen technique using s $s(n + 2)$ model assessments, Let $n$ is the number of input components and $s$ is the sample size. To address the problem, the starting sample size may be chosen, and the sensitivity indices can be determined. The sample size should then be increased, and the index computation should be performed again. The latest approved sample size is regarded acceptable when the derived indices stay unchanged, i.e., when they converge. $n = 6$ and $s = 1000$ points were used in this study. In Fig. 9 a, the findings of the Sobol'–Jansen based on Model I are shown. The influence of groundwater level and runoff flow on lake water level has been discovered. Figure 6 b shows how $\sigma_j$ and $\mu_j$ values can help distinguish between three types of input factors: (I) effective input factors with lowest non-linearity and interactive effects (high $\mu_j$ and low $\sigma_j$ values); (II) Influential input factors with non-linear and interactive effects (high $\mu_j$ and $\sigma_j$ values); and (III) input factors with lowest effect (low $\mu_j$ values).

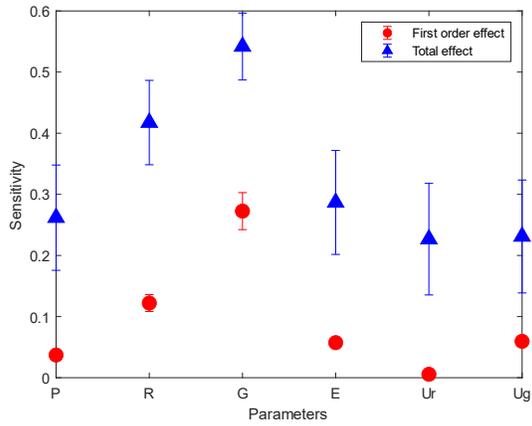
(a)
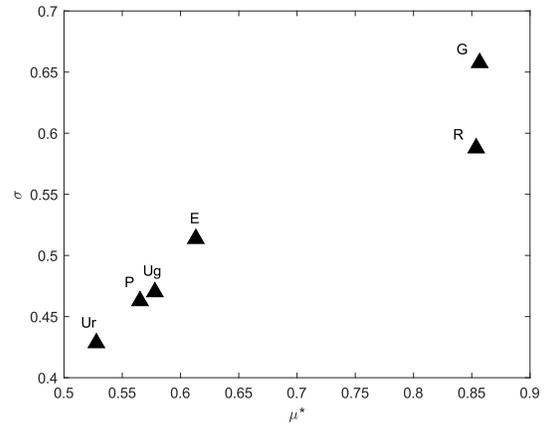
(b)
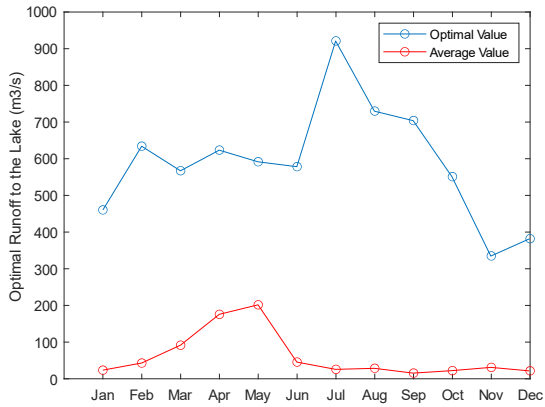
(c)
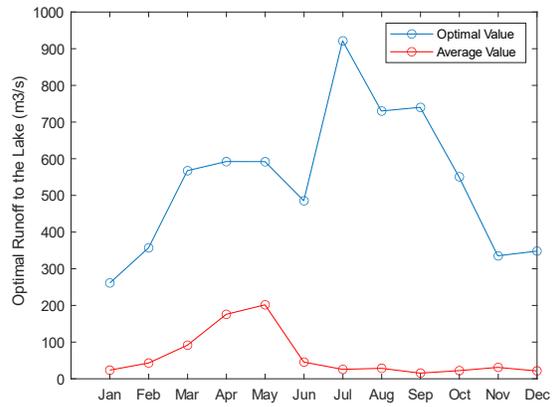
(d)
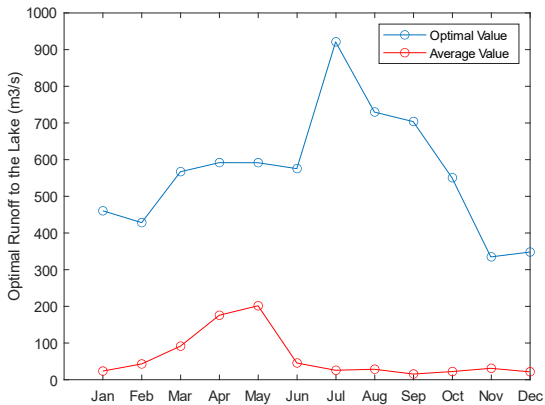
(e)
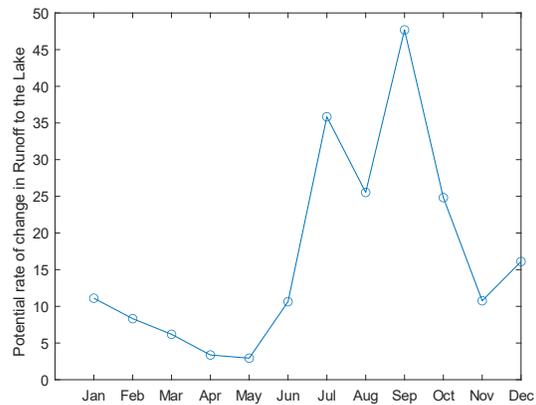
(f)

**Fig. 6** Results of sensitivity analysis using a) Sobol-Jansen method, b) Morris Method. Results of optimization using c) Genetic Algorithm, d) Nonlinear Optimization, e) Pattern search, and f) The potential rate of change in runoff rate to the Morris' findings corroborated those of the Sobol'–Jansen techniques. The groundwater level and runoff flow variables are particularly useful in this scenario. With non-linear and interactive effects, they have a high $\mu_j$ value and $\sigma_j$. Furthermore, evaporation is the third most important element affecting the lake's water level. Finally, the importance of Ug, P, and Ur is prioritized. According to the findings of the sensitivity analysis utilizing the Sobol'–Jansen and Morris technique, we can successfully regulate the lake water level by adjusting the ground water level and runoff water. We know that the groundwater level cannot be directly managed in the short term, thus environmental measures such as modifying agricultural patterns, optimizing urban area water assumptions, and other ways must be defined. We regard the groundwater level to be an uncontrolled variable in the short run in this scenario. However, long-term control will need more investigation. In the following part, we'll look at how rivers' runoff flows may be used to manage water levels.

### 4.4. Optimization of the Runoffs

In the next step, we provide Model II for the optimization problem after sensitivity analysis. Based on the sensitivity analysis results, we defined a new meta-model based on the controllable effective factor, River runoff, as a dependent variable. We try to maximize the river flow in the meta-model framework by fixing the lake water level based on the 2018 monthly pattern. The resulting runoff flow will be the potential R for monthly fixing of the lake water level. To evaluate the performance of the idea, we examined several optimization techniques and compared anticipated and optimal system values. Based on the proposed procedure, the objective function and constraints are defined as follows:

Maximize $R^* = f(H^*, P^*, G^*, E^*, Ur^*, Ug^*)$   s.t. $f: Model\ II$ (10)

$H^* = Hcon^*$ (11)

To solve the following optimization problem: Genetic algorithm, Non-linear optimization, and Pattern search using MATLAB 2020b optimization package. Regarding Eq. (10), the objective governing function is a non-linear and nonparametric equation that should be solved with non-linear methods. Moreover, the constraint of the lake level is defined as a constant value for each month based on the 2018 monthly pattern. The optimization problem is solved for each month using the presented methods.  Figure 6 shows the optimal value of total surface runoff to the lake. Regarding the results, R should be increased to a specific monthly rate to have a constant water level pattern. Regarding Figure 6, c, d, and e, the provided optimization algorithms illustrated almost similar results for the R. Regarding Figure 3, the lake level increased in November, December, January, February, March, April, May, and June. However, it decreases in July, August, September, and October. Suppose we classify this month as filling season and draining season. In that case, we can conclude that the surface runoff filling season should increase almost 8.7 times to have a constant pattern of the lake level. However, this value should be increased 33.5 times in the draining season. This plot can be a controlling pattern for the revitalization of the lake.

In this paper, optimal runoff water to the lake includes the total value of surface water and river discharges to the lake. Therefore, it is essential to distinguish between these two flows for controlling dam gates. Moreover, we presented a potential pattern for controlling inlet runoffs to the river in this case. Therefore, we do not use any information on dams' capacity, the number of rivers, and the flow rate of each river. Therefore, it is an exciting topic for future works to fill the control dams gate in a way that satisfies the presented pattern. Moreover, regarding the result of the sensitivity analysis, the most important factor in the lake's water level is the basin's groundwater level. To be concluded that there are central factors that are highly effective in the lake water level consist of groundwater level and runoff water to the lake. In this paper, we considered the groundwater level as uncontrollable parameters and concentrated on just surface runoffs to the river. However, groundwater levels are the most critical parameters. For future work, we suggest providing a new strategy for controlling the ground water level. The solution to the proposed problem directly affects the lake water level. The groundwater level also depends on several parameters such as agricultural pattern, drinking water consumption pattern, and perceptions that need a robust strategy to control this parameter.

## 5. Conclusion

This paper presents an optimization problem for determining how to stabilize Urmia Lake levels by finding the potential surface runoff flow or river discharge. Based on historical data, the presented procedure consists of two major stages: finding effective parameters in the lake water level based on sensitivity analysis approaches and optimizing the most influential variables in order to stabilize the lake water level with changing design variables. Our first step is to design a meta-model based on deep neural network algorithms in order to determine the variables that are effective. To determine the most effective input parameters, we used the Sobol-Jansen and Morris method. At each level, a local measure of incremental ratios, such as an elementary effect, is computed accounting for the entire input space. As a result, the variables are prioritized based on their effectiveness on the dependent variable. During the second stage, the input variables are arranged in a manner that makes the effective variables of the first stage dependent on the input variables of the second stage. Based on the results of the sensitivity analysis, surface runoff (R) is considered the output layer in the meta-model, while the remaining variables are considered the input layer. Consequently, the optimization is based on three methods: genetic algorithm, pattern search, and nonlinear programming.

Based on the monthly value of the evaporation from the lake, the maximum points of evaporation are in the summer season in Jun, July, and August. Moreover, precipitation has fluctuating values in both the spring and fall seasons. Despite evaporation with uniform variation, precipitation has a different pattern in different years. Similarly, total surface runoff flow to the river varies in different years. Regarding the river outflow to the lake, we can find a uniform pattern to control the water level in the lake. Furthermore, the ground water level at the end of the spring is at the maximum value, and the mid-fall has a minimum value. This variation refers to the end of the agricultural uses of the ground waters each year. Moreover, two yearly patterns are used as agricultural patterns of groundwater and river runoffs for agricultural uses. This pattern is extracted based on the results of the monthly time averaged Miyandoab aquifer between 2002 and 2013. Also, the maximum and minimum value of the lake water level belongs to June and November, respectively.

The provided meta-models are five-dimensional models with the best fitting of the historical data on the model. However, the model can estimate the dependent variable by changing other features. Building nonparametric models do not make explicit assumptions about the functional form, such as linear models in parametric models. In contrast, nonparametric modeling may be viewed as the closest function approximation to the datasets. By removing the presupposition of a certain functional form, such as a linear model, nonparametric models have the ability to properly match a greater range of possible real function forms. Therefore, we used the provided Model I as a nonparametric sensitivity analysis function. The results of the Sobol'–Jansen revealed that Groundwater level and Runoff flow have the most significant impacts on the lake water level. The Morris results also confirmed the findings of the Sobol'–Jansen methods. For this case, the Groundwater level and Runoff flow variable are highly effective with non-linear and interactive effects. Moreover, evaporation is the third most effective factor in the lake water level. Finally, agricultural use of groundwater, precipitation, and agricultural use of rivers are the next priority the importance. Based on the sensitivity analysis results using the Sobol'–Jansen and Morris method, it can be concluded that by changing the Groundwater level and Runoff water we can effectively control the lake water level. We used a Genetic algorithm, Non-linear optimization, and Pattern search to solve the following optimization problem.

Moreover, the constraint of the lake level is defined as a constant value for each month based on the 2018 monthly pattern. The optimization problem is solved for each month using the presented methods. We can conclude that the surface runoff filling season should increase almost 8.7 times to have a constant pattern of the lake level. However, this value should be increased 33.5 times in the draining season. This plot can be a controlling pattern for the revitalization of the lake. It is to be concluded that there are prominent factors that are highly effective in the lake water level consisting of groundwater level and runoff water to the lake. In this paper, we considered the groundwater level uncontrollable parameters and concentrated on just surface runoffs to the river. However, groundwater levels are the most critical parameters. For future work, we suggest providing a new strategy for controlling the groundwater level.

**Ethical Approval**

Not applicable

**Consent to Participate**

Not applicable

**Consent to Publish**

Not applicable

**Funding** The funding sources had no involvement in the study design, collection, analysis, or interpretation of data, writing of the manuscript, or in the decision to submit the manuscript for publication.

**Competing Interests**

We declare no conflict of interest.

**Availability of data and materials**

Based on the Urmia lake Restoration Program, all hydrometeorological data utilized in this work is available online: https://ndownloader.figstatic.com/articles/19207557/versions/2  Furthermore, other data is based on monthly time-averaged root-zone water balance for agricultural zones available online: https://ars.els-cdn.com/content/image/1-s2.0-S0378377419304068-mmc1.docx